\documentclass{article}

\usepackage{microtype,amssymb,hyperref,amsmath,color,amsthm}
\usepackage{graphicx}
\usepackage{subfigure}
\usepackage{booktabs} 

\newtheorem{thm}{Theorem}[section]
\newtheorem{prop}[thm]{Proposition}
\newtheorem{lemma}[thm]{Lemma}
\newtheorem{coro}[thm]{Corollary}
\newtheorem{exm}{Example}[section]
\newtheorem{rmk}{Remark}[section]

\allowdisplaybreaks

\usepackage{hyperref}

\usepackage[accepted]{icml2021}

\DeclareMathOperator*{\argmin}{argmin}
\DeclareMathOperator*{\argmax}{argmax}

\icmltitlerunning{Non-Exponentially Weighted Aggregation}

\begin{document}

\twocolumn[
\icmltitle{Non-Exponentially Weighted Aggregation:
\\
Regret Bounds for Unbounded Loss Functions}



\icmlsetsymbol{equal}{*}

\begin{icmlauthorlist}
\icmlauthor{Pierre Alquier}{pierre}
\end{icmlauthorlist}

\icmlaffiliation{pierre}{RIKEN AIP, Tokyo, Japan}

\icmlcorrespondingauthor{Pierre Alquier}{pierrealain.alquier@riken.jp}

\icmlkeywords{Aggregation of predictors, generalized Bayes, variational approximation, PAC-Bayes, regret bounds, $\phi$-divergences, online learning theory.}

\vskip 0.3in
]



\printAffiliationsAndNotice{}  

\begin{abstract}
We tackle the problem of online optimization with a general, possibly unbounded, loss function. It is well known that when the loss is bounded, the exponentially weighted aggregation strategy (EWA) leads to a regret in $\sqrt{T}$ after $T$ steps. In this paper, we study a generalized aggregation strategy, where the weights no longer depend exponentially on the losses. Our strategy is based on Follow The Regularized Leader (FTRL): we minimize the expected losses plus a regularizer, that is here a $\phi$-divergence. When the regularizer is the Kullback-Leibler divergence, we obtain EWA as a special case. Using alternative divergences enables unbounded losses, at the cost of a worst regret bound in some cases.
\end{abstract}

\section{Introduction}
\label{section:intro}

We focus in this paper on the online optimization problem as formalized for example in~\cite{shalev2012online}: at each time step $t\in\mathbb{N}$, a learning machine has to make a decision $\theta_t\in\Theta$. Then, a loss function $\ell_t:\Theta\rightarrow \mathbb{R}_+$ is revealed and the machine suffers loss $\ell_t(\theta_t)$. Typical example include online linear regression, where $\ell_t(\theta) = \left(y_t -\theta^T x_t\right)^2$ for some $x_t\in\mathbb{R}^d$ and $y_t\in\mathbb{R}$, or online linear classification with $\ell_t(\theta) = \mathbf{1}_{\{y_t\neq {\rm sign}( \theta^T x_t)\}}$ or $\ell_t(\theta) = \max(1-\theta^T x_t,0)$ for some $x_t\in\mathbb{R}^d$ and $y_t\in\{-1,+1\}$. The objective is to design a strategy for the machine that will ensure that the regret at time $T$,
\begin{equation}
\label{eqn:regret}
\mathcal{R}_T := \sum_{t=1}^T \ell_t(\theta_t) - \inf_{\theta\in\Theta} \sum_{t=1}^T \ell_t(\theta),
\end{equation}
satisfies $ \mathcal{R}_T = o(T)$.

Various strategies were investigated under different assumptions. When the functions $\ell_t$ are convex, methods based on the sub-gradient of $\ell_t$ can be used. Such strategies lead to regret in $\sqrt{T}$ under the additional assumption that the $\ell_t$ are Lipschitz. The regret bounds and strategies are detailed in Chapter 2 in~\cite{shalev2012online}. Another very popular strategy is the so-called exponentially weighted aggregation (EWA) that is based on the probability distribution:
\begin{equation}
\label{equa:bayes:closedform}
\rho^t({\rm d}\theta) = \frac{ \exp\left(-\eta \sum_{s=1}^{t-1} \ell_s(\theta) \right) \pi({\rm d}\theta) }
{\int \exp\left(-\eta \sum_{s=1}^{t-1} \ell_s(\vartheta) \right) \pi({\rm d} \vartheta)}
\end{equation}
for some prior distribution $\pi$ on $\Theta$ and some learning rate $\eta>0$. Drawing $\theta_t\sim\rho^t$ leads to an expected regret in $\sqrt{T}$, under the strong assumption that the losses $\ell_t$ are uniformly bounded, see~\cite{gerchinovitz2011prediction}.

Is is actually well known that
\begin{equation}
\label{equa:bayes:argmin}
\rho^t = \argmin_{\rho\in\mathcal{P}(\Theta)}
\left\{
\sum_{s=1}^{t-1}  \mathbb{E}_{\theta\sim \rho }[\ell_s(\theta)] + \frac{ {\rm KL}(\rho||\pi) }{\eta}
\right\}
\end{equation}
where ${\rm KL}$ is the Kullback-Leibler divergence and $\mathcal{P}(\Theta)$ is the set of all probability distributions on $\Theta$ equipped with a suitable $\sigma$-algebra (rigorous notations will come in Subsection~\ref{subsec:notations}). In this paper, we will study a generalization of the EWA strategy given by
\begin{equation}
\label{equa:bayes:argmin:general:divergence}
\rho^t = \argmin_{\rho\in\mathcal{P}(\Theta)}
\left\{
\sum_{s=1}^{t-1}  \mathbb{E}_{\theta\sim \rho }[\ell_s(\theta)] + \frac{ D_\phi(\rho||\pi) }{\eta}
\right\},
\end{equation}
where $D_\phi$ can be any $\phi$-divergence
(on the condition that this minimizer exists, which will be discussed). Such a strategy is known as ``Follow The Regularized Leader'' (FTRL) in the online optimization community, and has been studied extensively in the finite $\Theta$ case~\cite{shalev2012online,HazanOnlineConvexOptimization,orabona2019modern}. In Bayesian statistics,~\eqref{equa:bayes:argmin:general:divergence} was advocated recently in~\cite{li2016renyi,knoblauch2019generalized}. Some generalization error bounds were proven by~\cite{alquier2018simpler}. However,~\cite{alquier2018simpler} is written in the batch setting, and the error bounds thus require strong assumptions: for $\theta\in\Theta$, the $(\ell_t(\theta))_{t\in\mathbb{N}}$ must be independent and identically distributed random variables.

Let us call $\rho^t$ the $D_\phi$-posterior associated to the $\phi$-divergence $D_\phi$, the prior $\pi$, the learning rate $\eta$ and the sequence of losses $(\ell_s)_{s\in\mathbb{N}}$. In this paper, we study $D_\phi$-posteriors in the online setting, which allows to get completely rid of the stochastic assumptions of~\cite{alquier2018simpler}. First, we prove a regret bound on the $D_\phi$-posterior. Our proof follows the same scheme as the study of FTRL in~\cite{shalev2012online,orabona2019modern}, but in the general case ($\Theta$ is not assumed to be finite). Interestingly, when $D_\phi$ is the $\chi^2$ divergence, our bound holds under very general assumptions -- in particular, it does not require that the losses are bounded, Lipschitz, nor convex, but that might be at the cost of a larger regret. We also provide explicit forms for the $D_\phi$-posterior. It turns out that it extends the idea of EWA beyond the exponential function, thus the title of the paper. Finally, it is known that EWA is not always feasible in practice. A way to overcome this issue is to use variational approximations of EWA. We thus propose an algorithm that can be seen as the generalization of online variational inference to $\phi$-divergences, and provide a regret bound.

\subsection{Related works}

The case $D_\phi = {\rm KL}$,~\eqref{equa:bayes:argmin} has been studied under the name ``multiplicative update'', aggregating strategy, EWA~\cite{Vovk:1990:AS:92571.92672,littlestone1994weighted,catoni2004statistical} to name a few. Regret bounds in $\sqrt{T}$ can be found in~\cite{stoltz2005incomplete,cesa2006prediction,devaine2013forecasting} in the case where $\Theta$ is finite, we refer the reader to~\cite{gerchinovitz2011prediction} for the general case. Note that in~\cite{shalev2012online,HazanOnlineConvexOptimization,orabona2019modern}, EWA is studied as a special case of the FTRL strategy (Follow The Regularized Leader), in the case where $\Theta$ is finite. This point of view is the main inspiration of the proofs in this paper, even though we deal here with a general set $\Theta$. Also, note that smaller regret in $\log T$ is feasible under a stronger assumption: exp-concavity~\cite{HazanLogarithmic,cesa2006prediction,audibert2009fast}. \cite{reid2015generalized,mham2018constant} also studied small regrets and used for this a generalization of EWA beyond the KL divergence, but here again the study was restricted to a finite set $\Theta$. Similar techniques were also considered by~\cite{audibert2009minimax,pmlr-v89-zimmert19a} in incomplete information problems (bandits).

Given a statistical model, that is, a family of densities $p_\theta$ with respect to a reference measure $\nu$ on some space $\mathcal{X}$, and i.i.d random variables $X_1$, $X_2$, $\dots$, drawn from some probability distribution on $\mathcal{X}$, one can define the loss $\ell_t(\theta) = -\log p_\theta (X_t) $. In this case, for $\eta=1$, $\rho^t$ is actually the posterior distribution of $\theta$ given $X_1,\dots,X_{t-1}$ used in Bayesian statistics. Thus, EWA is also sometimes refered to as ``generalized Bayes''. \cite{li2016renyi} proposed~\eqref{equa:bayes:argmin:general:divergence} as one further generalization of Bayes, using R\'enyi divergences instead of ${\rm KL}$. More recently,~\cite{knoblauch2019generalized} advocated for a use of taylored losses and divergences. Note that in the batch setting, a general theory allows to provide risk bounds for generalized Bayes (or EWA): PAC-Bayes bounds~\cite{shawe1997PAC,mcallester1999some,MR2483528,alquier2008pac}, see~\cite{guedj2019primer} for a recent survey. PAC-Bayes bounds for generalized Bayes with the $\chi^2$-divergence were proven in~\cite{honorio2014tight} and for the R\'enyi divergence in~\cite{begin2016pac}. \cite{alquier2018simpler,ohnishi2021novel} showed that while these bounds are usually less tight than standard PAC-Bayes bounds, they allow to get rid of the boundedness assumption in these results. The corresponding optimal posteriors are derived in~\cite{alquier2018simpler}. Other techniques to get rid of boundedness are discussed in~\cite{holland2019pac,rivasplata2020pac} in the batch case.

The idea of variational approximations is to minimize~\eqref{equa:bayes:argmin} over a restricted set of probability distributions in order to get a feasible approximation of $\rho^t$, see~\cite{blei2017variational,alquier2020approximate} for recent surveys. In the online setting, online variational approximations are studied by~\cite{khan2017conjugate,khan2018} and led to the first scaling of Bayesian principles to state-of-the-art neural networks~\cite{osawa2019practical}. In the i.i.d setting, a series of paper established the first theoretical results on variational inference, for many of them through a connection with PAC-Bayes bounds~\cite{alquier2016properties,NIPS2017_7100,dziugaite2018entropy,cherief2018consistency,cherief2019DEEP,wang2019frequentist,Tempered,yang2020alpha,wang2019variational,jaiswal2019asymptotic,Chicago,cherief2020contributions,plummer2020dynamics,ban2021markov,frazier2021loss,medina2021robustness}. Up to our knowledge, the only regret bound for online variational inference can be found in~\cite{cherief2019generalization}. The analysis of our generalized online variational approximation is based on this work.

\subsection{Notations}
\label{subsec:notations}

Let us now provide accurate notations and a few basic assumptions that will be used throughout the paper. We assume that the set $\Theta$ is equipped with a $\sigma$-algebra $\mathcal{T}$. Let $\mathcal{P}(\Theta)$ denote the set of all probability distributions on $(\Theta,\mathcal{T})$. Let $\pi\in\mathcal{P}(\Theta)$ be a probability distribution called the {\it prior} and $(\ell_s)_{s\in\mathbb{N}}$ be a sequence of functions called losses, $\ell_s:\Theta\rightarrow \mathbb{R}_+$, assumed to be $\mathcal{T}$-measurable.

Let $\mathcal{M}(\Theta)$ be the set of all finite, signed measures on $(\Theta,\mathcal{T})$. Note that $\mathcal{P}(\Theta) \subsetneq \mathcal{M}(\Theta)$. A norm $N$ on $ \mathcal{M}(\Theta)$ is a function $N:\mathcal{M}(\Theta) \rightarrow [0,\infty]$ with i) $N(\nu)=0 \Leftrightarrow \nu = 0$, ii) $N(\nu+\mu) \leq N(\nu) + N(\mu)$ and iii) for $\lambda\in\mathbb{R}$, $N(\lambda.\nu)= |\lambda| N(\nu)$. A norm $N$ on $\mathcal{M}(\Theta)$ induces a metric on $\mathcal{P}(\Theta)$ given by $d_N(\mu,\nu) = N(\nu-\mu)$. For example, the total variation norm $N_{{\rm TV}}(\nu) = \sup_{A \in \mathcal{T}}|\nu(A)|$ leads to the classical total variation distance on $\mathcal{P}(\Theta)$.

Given a sctrictly convex function $\phi: \mathbb{R}_+ \rightarrow \mathbb{R}\cup\{+\infty\}$ with $\phi(1)=0$ and $ \inf_{x\geq 0}\phi(x)  >-\infty$, define the $\phi$-divergence between $\rho$ and $\pi\in \mathcal{P}(\Theta)$ by
\begin{equation}
\label{dfn:Phi:divergence}
D_\phi(\rho||\pi) =
\mathbb{E}_{\theta\sim\pi} \left[ \phi\left(\frac{{\rm d}\rho}{{\rm d}\pi}(\theta) \right) \right] \text{ if } \rho \ll \pi
\end{equation}
and $+\infty$ otherwise. By Jensen's inequality, $D_\phi(\rho||\pi)\geq 0$. Put $\mathcal{P}_{D_\phi,\pi}(\Theta)=\{\rho\in\mathcal{P}(\Theta):D_\phi(\rho||\pi)<+\infty\}$.

A real-valued function $f$ is said to be upper semicontinuous if for any $\alpha$, $\{x: f(x)\geq \alpha \}$ is closed. For any real-valued function $f$, we will denote by $f_+$ the function defined by $f_+(x) = \max(f(x),0)$. Given a function $f:\mathbb{R}^d \rightarrow \mathbb{R}\cup\{+\infty\}$ that is not uniformly infinite, we will let $f^*$ denote its convex conjugate, that is, for any $y \in \mathbb{R}^d$,
\begin{equation}
\label{dfn:convex:conjugate:general}
f^*(y) =
\sup_{x\in\mathbb{R}^d} \left\{ x^Ty-f(x) \right\} \in \mathbb{R}\cup \{+\infty\}.
\end{equation}

\subsection{Outline of the paper}

We state our general regret bound in Section~\ref{section:regret}. In particular, we show that for some divergences, our result extends the results known for EWA to unbounded losses. We then provide an explicit form for the $D_\phi$-posterior in Section~\ref{section:explicit:posterior}. We study generalized online variational inference in Section~\ref{section:approximate}. Section~\ref{section:proofs} contains the proofs of the results in Sections~\ref{section:regret} and~\ref{section:explicit:posterior}, the remaining proofs are in the Appendix.

\section{A Regret Bound for $D_\phi$-Posteriors}
\label{section:regret}

\subsection{General result}

\begin{thm}
\label{thm:regret_bound}
Assume that there is a norm $N$ on $\mathcal{M}(\Theta)$ and real numbers $\alpha,L>0$ such that
\begin{itemize}
 \item for any $\rho \in\mathcal{M}(\Theta)$, $N(\rho) \geq N_{{\rm TV}}(\rho) $,
 \item for any $t\in\mathbb{N}$, for any $(\rho,\rho')\in\mathcal{P}_{D_\phi,\pi}(\Theta)^2$,
 \begin{equation}
 \label{asm:Lip:ell}
  \left|\mathbb{E}_{\theta\sim\rho} [\ell_t(\theta)] - \mathbb{E}_{\theta\sim\rho'} [\ell_t(\theta)] \right| \leq L N(\rho-\rho'),
 \end{equation}
 \item  for any $\gamma\in[0,1]$, for any $(\rho,\rho')\in\mathcal{P}_{D_\phi,\pi}(\Theta)^2$,
 \begin{multline}
  \label{asm:strCvx:D}
  D_\phi(\gamma \rho + (1-\gamma)\rho '||\pi) \leq - 2 \alpha \gamma (1-\gamma) N(\rho-\rho')^2
  \\ + \gamma D_\phi(\rho||\pi) + (1-\gamma)D_\phi(\rho '||\pi)  .
\end{multline}
\end{itemize}
Assume that each $\ell_t$ is $\pi$-integrable.
Then $\rho^t$ in~\eqref{equa:bayes:argmin:general:divergence} exists, is unique, and
 \begin{multline}
  \label{equa:thm:regret}
 \sum_{t=1}^T \mathbb{E}_{\theta\sim \rho^t}[\ell_t(\theta)]
 \leq \inf_{\rho\in\mathcal{P}(\Theta)}
 \Biggl\{
 \sum_{t=1}^T \mathbb{E}_{\theta \sim \rho}[\ell_t(\theta)]
 \\
 + \frac{\eta L^2 T}{\alpha} + \frac{ D_\phi(\rho||\pi) }{\eta}
 \Biggr\}.
 \end{multline}
\end{thm}

The assumptions have a simple interpretation:~\eqref{asm:Lip:ell} states that each $\mathbb{E}_{\theta\sim\rho}[\ell_t]$ is $L$-Lipschitz in $\rho$ with respect to the norm $N$, while~\eqref{asm:strCvx:D} states that $D_\phi$, as a function of its first argument, is $\alpha$-strongly convex with respect to $N$.

Regarding the choice of $\eta$, $\eta\sim 1/\sqrt{T}$ seems natural (indeed, in the countable case studied below, it leads to regrets in $\sqrt{T}=o(T)$). However, this choice depends on the horizon $T$. The doubling trick can be used to avoid this dependence, see e.g.~\cite{cesa2006prediction}.

When the losses $\ell_t$ are convex, Jensen's inequality gives $\mathbb{E}_{\theta\sim \rho^t}[\ell_t(\theta)] \geq \ell_t[\mathbb{E}_{\theta\sim \rho^t}(\theta)]$. We can thus use the posterior mean $\mathbb{E}_{\theta\sim \rho^t}(\theta)$ instead of a randomized strategy.

\begin{coro}
 Under the assumptions of Theorem~\ref{thm:regret_bound}, assuming moreover that each $\ell_t$ is convex, and writing $\hat{\theta}_t = \mathbb{E}_{\theta\sim \rho^t}(\theta)$, we have
  \begin{multline}
 \sum_{t=1}^T \ell_t(\hat{\theta}_t)
 \leq \inf_{\rho\in\mathcal{P}(\Theta)}
 \Biggl\{
 \sum_{t=1}^T \mathbb{E}_{\theta \sim \rho}[\ell_t(\theta)]
 \\
 + \frac{\eta L^2 T}{\alpha} + \frac{ D_\phi(\rho||\pi) }{\eta}
 \Biggr\}.
 \end{multline}
\end{coro}
We now apply Theorem~\ref{thm:regret_bound} to classical divergences.
\begin{exm}
Consider $\phi(x) = x\log x$ so that $D_\phi(\rho||\pi) = {\rm KL}(\rho||\pi)$ the Kullback-Leibler divergence. Assuming that, for any $t\in\mathbb{N}$, $|\ell_t(\theta)| \leq L$ holds $\pi$-almost surely on $\theta$, we have, for $(\rho,\rho')\in\mathcal{P}_{D_\phi,\pi}(\Theta)^2$,
\begin{align}
\int & \ell_t(\theta) \rho({\rm d}\theta) - \int \ell_t \rho'({\rm d}\theta)
\\
& = \int \ell_t(\theta) \left| \frac{{\rm d}\rho}{{\rm d}\pi}(\theta)-\frac{{\rm d}\rho'}{{\rm d}\pi}(\theta)\right| \pi({\rm d}\theta)
\\
& \leq L \int  \left| \frac{{\rm d}\rho}{{\rm d}\pi}(\theta)-\frac{{\rm d}\rho'}{{\rm d}\pi}(\theta)\right| \pi({\rm d}\theta)
\end{align}
that is,~\eqref{asm:Lip:ell} holds with the norm on $\mathcal{M}(\Theta)$:
\begin{align}
 N(\rho) = \int \left| \frac{{\rm d}\rho}{{\rm d}\pi}(\theta)\right| \pi({\rm d}\theta) = 2 N_{{\rm TV}}(\rho).
\end{align}
It is known that~\eqref{asm:strCvx:D} holds with $\alpha=1$, the calculations are detailed in the discrete case page 30 in~\cite{shalev2012online} and can be directly extended to the general case. So
 \begin{multline}
 \sum_{t=1}^T \mathbb{E}_{\theta\sim \rho^t}[\ell_t(\theta)]
 \leq \inf_{\rho\in\mathcal{P}(\Theta)}
 \Biggl\{
 \sum_{t=1}^T \mathbb{E}_{\theta \sim \rho}[\ell_t(\theta)]
 \\
 + \eta L^2 T + \frac{ {\rm KL}(\rho||\pi) }{\eta}
 \Biggr\}.
 \end{multline}
This is essentially the same result as Theorem 2.2 page 16 in~\cite{cesa2006prediction}. Note however that a different proof technique is used there, that leads to better constants: the term in $\eta L^2 T$ is replaced by $\eta L^2 T/8$.
\end{exm}

Before considering a new example, let us simply remind the definition of strong convexity: a function $\varphi:\mathbb{R}^d \rightarrow \mathbb{R}$ is said to be $\alpha$-strongly convex with respect to a norm $\|\cdot\|$ when, for any $(u,v)\in(\mathbb{R}^d)^2$ and $\gamma\in[0,1]$,
$  \varphi(\gamma u + (1-\gamma)v) \leq \gamma \varphi(u) + (1-\gamma) \varphi(v)  - \alpha \gamma(1-\gamma)\|u-v\|^2 /2$.
It is known that when $\varphi:\mathbb{R}\rightarrow \mathbb{R}$ is twice differentiable and $\|\cdot\|$ is the Euclidean norm, this is equivalent to the condition: $\forall u\text{, }\varphi''(u) \geq \alpha$. Plugging $u=\frac{{\rm d}\rho}{{\rm d}\pi}(\theta)$ and $v=\frac{{\rm d}\rho'}{{\rm d}\pi}(\theta)$ in this definition and integrating with respect to $\pi$ immediately yields the following.

\begin{lemma}
\label{lemma:strCvx}
Assume that $\phi:\mathbb{R} \rightarrow \mathbb{R}$ with $\phi(1)=0$ is $\alpha$-strongly convex, then the $\phi$-divergence $D_\phi$ satisfies~\eqref{asm:strCvx:D} for the $\mathcal{L}_2(\pi)$-norm
\begin{equation}
 N_2(\rho) := \sqrt{\int \left(\frac{{\rm d}\rho}{{\rm d}\pi}(\theta)  \right)^2 \pi({\rm d}\theta)} \geq 2 N_{{\rm TV}}(\rho)
\end{equation}
(extended by $+\infty$ when $\rho \ll \pi$ does not hold).
\end{lemma}

\begin{exm}
Now, $\phi(x) = x^2-1$, so $D_\phi(\rho||\pi) = \chi^2(\rho||\pi)$ the $\chi^2$-divergence. As $x\mapsto x^2$ is $2$-strongly convex, Lemma~\ref{lemma:strCvx} gives~\eqref{asm:strCvx:D} with $N=N_2$. Moreover,
\begin{align}
 \Biggl|\int & \ell_t(\theta) \rho({\rm d}\theta) - \int \ell_t \rho'({\rm d}\theta)\Biggr|
\nonumber
\\
& \leq \int \ell_t(\theta) \left| \frac{{\rm d}\rho}{{\rm d}\pi}(\theta)-\frac{{\rm d}\rho'}{{\rm d}\pi}(\theta)\right| \pi({\rm d}\theta)
\nonumber
\\
& \leq N_2(\rho-\rho') \left(\int \ell_t(\theta)^2  \pi({\rm d}\theta)\right)^{1/2}.
\end{align}
So,  we obtain~\eqref{asm:Lip:ell} under the only assumption that, for any $t\in\mathbb{R}$, $
 \int \ell_t(\theta)^2  \pi({\rm d}\theta) \leq L^2
$.
\end{exm}
As the application of Theorem~\ref{thm:regret_bound} to the context of the previous example is new to our knowledge, we state it now as a separate corollary.
\begin{coro}
\label{coro:regret:chi2}
Define $\rho^t$ as in~\eqref{equa:bayes:argmin:general:divergence} with $D_\phi=\chi^2$. Assume that for any $t\in\mathbb{R}$,
\begin{equation}
\label{equa:assumption:L2}
\int \ell_t(\theta)^2  \pi({\rm d}\theta) \leq L^2
\end{equation}
for some $L>0$, then
 \begin{multline}
 \sum_{t=1}^T \mathbb{E}_{\theta\sim \rho^t}[\ell_t(\theta)]
 \leq \inf_{\rho\in\mathcal{P}(\Theta)}
 \Biggl\{
 \sum_{t=1}^T \mathbb{E}_{\theta \sim \rho}[\ell_t(\theta)]
 \\
 + \frac{\eta L^2 T}{2} + \frac{ \chi^2(\rho||\pi) }{\eta}
 \Biggr\}.
 \end{multline}
\end{coro}
It is important to note that~\eqref{equa:assumption:L2} allows choices of priors that are not possible with EWA. Consider for example classification with the exponential loss $\ell_t(\theta)=\exp(-y_t x_t^T\theta)$ or with the hinge loss $\ell_t(\theta)=\max(0,1-y_t x_t^T\theta)$, where $\Theta = \mathbb{R}^d$, $y_t\in\{-1,+1\}$ and $\|x_t\|\leq 1$. In this case,~\eqref{equa:assumption:L2} will be satisfied with any Gaussian prior. However, we don't have $\ell_t(\theta)\leq L$ uniformly on $\mathbb{R}^d$: this prevents to use EWA with such a prior. Another example (quadratic loss) is provided in Appendix A.

\begin{rmk}
 One of the anonymous Referees suggested an alternative proof for Corollary~\ref{coro:regret:chi2}, in the finite $\Theta$ case: rewrite the $\chi^2$ divergence as a weighted quadratic norm between $\rho$ and $\pi$, and use the results on weighted $\ell_p$ norms in Section 5 of~\cite{orabona2015generalized}. This would require some adaptation of the proof to constrain $\rho$ to belong to the simplex, but it would be interesting to compare Corollary~\ref{coro:regret:chi2} to the results obtained in this way.
\end{rmk}

\subsection{Comparison of the bounds in the countable case}

In this subsection, $\Theta = \{\theta_0,\theta_1,\dots\}$ is countable. Consider any prior $\pi$. In this case, we upper bound the infimum in~\eqref{equa:thm:regret} by its restriction to all Dirac masses. We obtain:
\begin{coro}
 Under the conditions of Theorem~\ref{thm:regret_bound}, assuming in addition that $\Theta = \{\theta_0,\theta_1,\dots\}$ we have:
  \begin{multline}
  \label{equa:thm:regret:2}
 \sum_{t=1}^T \mathbb{E}_{\theta\sim \rho^t}[\ell_t(\theta)]
 \leq \inf_{j\in\mathbb{N}}
 \Biggl\{
 \sum_{t=1}^T \ell_t(\theta_j)
 + \frac{\eta L^2 T}{\alpha}
 \\
 +\frac{ \pi(\theta_j) \phi\left(\frac{1}{\pi(\theta_j)}\right) +(1-\pi(\theta_j))\phi(0)  }{\eta}
 \Biggr\}.
 \end{multline}
\end{coro}
In any case, chosing $\eta=1/\sqrt{T}$ will lead to a regret in $\sqrt{T}$:
  \begin{multline}
 \sum_{t=1}^T \mathbb{E}_{\theta\sim \rho^t}[\ell_t(\theta)]
 \leq \inf_{j\in\mathbb{N}}
 \Biggl\{
 \sum_{t=1}^T \ell_t(\theta_j)
 +  \Biggl[ \frac{L^2}{\alpha}
 \\
 \pi(\theta_j) \phi\left(\frac{1}{\pi(\theta_j)}\right) +(1-\pi(\theta_j))\phi(0)
 \Biggr] \sqrt{T} \Biggr\}.
 \end{multline}
Regarding the dependence on $\pi$, let us now to compare the bounds for $D_\phi={\rm KL}$ and $D_\phi=\chi^2$.
\begin{exm}
When $D_\phi={\rm KL}$, the assumption in~\eqref{asm:Lip:ell} implies that $0\leq \ell_t(\theta_j) \leq L$ for any $t,j\in\mathbb{N}$. In the case $\ell_t(\theta) = |y_t - f_{\theta}(x_t) |$ this can be obtained by assuming that $|y_t|\leq L/2$ where $L$ is known, so that the predictors will be designed or truncated by the user to stay in the interval $[-L/2,L/2]$.
In this case, the bound in~\eqref{equa:thm:regret:2} becomes
  \begin{multline}
 \sum_{t=1}^T \mathbb{E}_{\theta\sim \rho^t}[\ell_t(\theta)]
 \leq \inf_{j\in\mathbb{N}}
 \Biggl\{
 \sum_{t=1}^T \ell_t(\theta_j)
 + \eta L^2 T
 \\
 + \frac{  \log\left(\frac{1}{\pi(\theta_j)} \right) }{\eta}
 \Biggr\}.
 \end{multline}
\end{exm}

\begin{exm}
 When $D_\phi=\chi^2$, the bound in~\eqref{equa:thm:regret:2} becomes
  \begin{multline}
 \sum_{t=1}^T \mathbb{E}_{\theta\sim \rho^t}[\ell_t(\theta)]
 \leq \inf_{j\in\mathbb{N}}
 \Biggl\{
 \sum_{t=1}^T \ell_t(\theta_j)
 + \frac{\eta L^2 T}{2}
 \\
 + \frac{  \frac{1}{\pi(\theta_j)} -1 }{\eta}
 \Biggr\}
 \end{multline}
 which is much worse for large $j$'s (for which we necessarily will have $\pi(\theta_j)$ small). On the other hand, the assumption in~\eqref{asm:Lip:ell} only requires
 \begin{equation}
0\leq \sum_{j=0}^{\infty} \pi(\theta_j) \ell_t(\theta_j)^2 \leq L^2
 \end{equation}
 for any $t\in\mathbb{N}$. In the case $\ell_t(\theta) = |y_t - f_{\theta}(x_t) |$ this can be obtained by assuming that $|y_t|\leq c$ where $c$ is unknown. Indeed the user might be tempted to use predictors with various magnitude: $|f_{\theta_j}(x)|\leq c_j$ where $c_j$ grows with $j$. In order to ensure~\eqref{asm:Lip:ell} we must take a prior $\pi$ such that
 \begin{equation}
  L^2 := 2c^2 + 2 \sum_{j=0}^\infty \pi(\theta_j) c_j^2 < +\infty.
 \end{equation}
\end{exm}
\begin{rmk}
 The take-home message of these examples is that the $\chi^2$ divergence allows unbounded losses, but at the cost of a worst regret bound. One of the anonymous Referees asked whether it is possible to get the best of both worlds, that is, unbounded losses with the same regret bound of EWA. This is of course a very important question, we are not aware of existing answers. In an additional example in Appendix A, we show however that it is possible to mitigate the deterioration of the bound in the unbounded case.
\end{rmk}
 \begin{rmk}
 Another anonymous Referee pointed out that~\cite{kalnishkan2008weak} also derived regret bounds for unbounded losses. In their bound~(8), there is a term in $\varepsilon T$ where $\varepsilon>0$ is some tuning parameter, thus, when $\epsilon$ is constant, their bound is in $T$ and not in $\sqrt{T}$. Chosing $\varepsilon=1/\sqrt{T}$ in their bound leads to non-explicit regret bounds because of the term $L_\varepsilon$.
\end{rmk}

\section{Explicit $D_\phi$-Posteriors: Non-Exponentially Weighted Aggregation}
\label{section:explicit:posterior}

We now provide an explicit formula for the $D_\phi$-posterior $\rho^t$.

\begin{prop}
 \label{prop:posterior:general:closedform}
 Assume that $\phi$ is differentiable, strictly convex and define $\tilde{\phi}$ on $\mathbb{R}$ by $\tilde{\phi}(x)=\phi(x)$ if $x\geq 0$ and $\tilde{\phi}(x)=+\infty$ otherwise. Then
 \begin{equation}
 \label{equa:prop:tilde}
  \tilde{\phi}^* = \sup_{x\in\mathbb{R}}[xy-\tilde{\phi}(x)] = \sup_{x\geq 0}[xy-\phi(x)]
 \end{equation}
is differentiable and for any $y\in\mathbb{R}$,
\begin{equation}
 \label{equa:posterior:general:differential}
 \nabla \tilde{\phi}^* (y) = \argmax_{x\geq 0} \left\{ xy-\phi(x) \right\}.
\end{equation}
Assume moreover that $\tilde{\phi}^*(\lambda-a)-\lambda \rightarrow \infty$ when $\lambda \rightarrow \infty$, for any $a\geq 0$. Then
 \begin{equation}
\label{equa:posterior:general:constant:closedform}
\lambda_t = \argmin_{\lambda\in\mathbb{R}} \left\{ \int \tilde{\phi}^* \left(\lambda-\eta \sum_{s=1}^{t-1} \ell_s(\theta) \right) \pi({\rm d}\theta) - \lambda \right\}
\end{equation}
exists, and
 \begin{equation}
 \label{equa:posterior:general:closedform}
\rho^t({\rm d}\theta) = \nabla \tilde{\phi}^* \left(\lambda_t-\eta\sum_{s=1}^{t-1}\ell_s(\theta)\right) \pi({\rm d}\theta)
\end{equation}
minimizes~\eqref{equa:bayes:argmin:general:divergence}.
\end{prop}
In the finite $\Theta$ case,~\eqref{equa:posterior:general:closedform} was proven by~\cite{reid2015generalized}. An anonymous Referee pointed out that it can also be recovered thanks to~\cite{teboulle1992entropic}. The techniques used in these papers cannot be used in the general case, though. Instead, we use new tools from~\cite{agrawal2020optimal}, that are introduced in the proof.

A similar formula in the context of bandits (with a finite number of arms) can also be found in~\cite{audibert2009minimax}. The distribution $\rho^t$ is also related to the generalized exponential family in~\cite{grunwald2004game} and the generalized MaxEnt models of~\cite{frongillo2014convex}.

\begin{exm}
First, $\phi(x) = x\log(x)$ so $D_\phi = {\rm KL}$. In this case, $\tilde{\phi}^*(y) = \exp(y-1)$ so $\nabla \tilde{\phi}^*(y) = \tilde{\phi}^*(y) = \exp(y-1)$. This leads to
 \begin{equation}
\label{equa:posterior:general:constant:closedform:KL}
\lambda_t = - \log \int \exp \left[-\eta \sum_{s=1}^{t-1} \ell_s(\theta)-1 \right] \pi({\rm d}\theta),
\end{equation}
and
 \begin{equation}
 \label{equa:posterior:general:closedform:KL}
\rho^t({\rm d}\theta)
= \frac{ \exp \left[-\eta \sum_{s=1}^{t-1} \ell_s(\theta) \right] \pi({\rm d}\theta)}
{\int \exp \left[-\eta \sum_{s=1}^{t-1} \ell_s(\vartheta) \right] \pi({\rm d}\vartheta)}.
\end{equation}
 
\end{exm}

\begin{exm}
Then $\phi(x) = x^2-1$, so $D_\phi = \chi^2$. In this case, $\tilde{\phi}^*(y) = (y^2/4)\mathbf{1}_{\{y\geq 0\}} $, so $\nabla \tilde{\phi}^*(y) = (y/2)_+ $ and
 \begin{equation}
 \label{equa:posterior:general:closedform:chi2}
\rho^t({\rm d}\theta)
= \left[\frac{\lambda_t -\eta \sum_{s=1}^{t-1} \ell_s(\theta)}{2} \right]_+ \pi({\rm d}\theta).
\end{equation}
In this case, $\lambda_t$ is not available in closed form, but it exists and is the only constant that will make the above sum to $1$.
\end{exm}

\begin{exm}
 More generally, consider $\phi(x) = x^p-1$. In this case $\nabla \tilde{\phi}^*(y)=(y/p)_+^{1/(p-1)} $, which leads to
 \begin{equation}
 \label{equa:posterior:general:closedform:chip}
\rho^t({\rm d}\theta)
= \left[\frac{\lambda_t -\eta \sum_{s=1}^{t-1} \ell_s(\theta)}{p} \right]_+ ^{\frac{1}{p-1}}\pi({\rm d}\theta).
\end{equation}
This is quite similar to the Polynomially Weighted Average forecaster studied in Corollary 2.1 page 12 in~\cite{cesa2006prediction} in the finite $\Theta$ case, even though the normalization procedure is different.
\end{exm}

\begin{rmk}
\label{rmk:posterior}
 When $\Theta=\{\theta_1,\dots,\theta_M \}$ is finite, these results are simply opbtained by minimizing
 \begin{equation}
 F(\rho^t_1,\dots,\rho^t_M) =  \sum_{j=1}^M \rho^t_j \sum_{s=1}^{t-1} \ell_s(\theta_j) + \frac{\pi_j \phi \left(\frac{\rho^t_j}{\pi_j}\right) }{\eta}
 \end{equation}
 under the constraint that $\rho^t_1+\dots+\rho^t_M =1$ and that for all $j$, $\rho_j\geq 0$ (for the sake of simplicity, we wrote $\pi_j:= \pi(\theta_j)$ and $\rho^t_j:= \rho^t(\theta_j)$). The Lagrange operator is given by
 \begin{multline}
 \mathcal{L}(\rho^t_1,\dots,\rho^t_M,\lambda,\nu_1,\dots,\nu_M) 
 =  \sum_{j=1}^M \rho^t_j \sum_{s=1}^{t-1} \ell_s(\theta_j) 
 \\
 + \frac{\sum_{j=1}^M \pi_j \phi \left(\frac{\rho^t_j}{\pi_j}\right) }{\eta} + \lambda \frac{ 1-\sum_{j=1}^M \rho^t_j}{\eta} + \sum_{j=1}^M \nu_j \rho^t_j
 \end{multline}
 (the notation $\lambda$ is carefully chosen: it indeed corresponds to~\eqref{equa:posterior:general:constant:closedform}).
Under the assumptions of Proposition~\ref{prop:posterior:general:closedform}, the method of Lagrange multipliers will lead to~\eqref{equa:posterior:general:differential}. We believe that this derivation gives some insights on~\eqref{equa:posterior:general:closedform}. So, we provide it in full length in Appendix C.
\end{rmk}

\section{Generalized Online Variational Inference}
\label{section:approximate}

Apart from the special case of conjugacy, the probability distribution $\rho^t$ in~\eqref{equa:bayes:closedform} is not tractable. Thus, $\rho^t$ in~\eqref{equa:bayes:argmin:general:divergence} is not expected to be tractable either. It can of course be implemented via Monte-Carlo methods, but the cost of these methods is often prohibitive for the online setting. In~\cite{cherief2019generalization}, the authors proposed to use a variational approximation, that is, to minimize~\eqref{equa:bayes:argmin} on a set smaller than $\mathcal{P}(\Theta)$. We here propose to extend this idea to the minimization in~\eqref{equa:bayes:argmin:general:divergence}.

\subsection{The algorithm}

Let $(q_\mu)_{\mu\in M}$ be a set of probability distributions in $\mathcal{P}(\Theta)$, where $M$ is some closed convex set in $\mathbb{R}^d$. We could define the variational approximation of $\rho^t$ in this family by:
\begin{equation}
\label{equa:approximate:argmin:constaint}
\argmin_{\mu\in M}
\left\{
\sum_{s=1}^{t-1} \mathbb{E}_{\theta\sim q_{\mu} }[\ell_s(\theta)] + \frac{ D_\phi(q_\mu||\pi) }{\eta}
\right\},
\end{equation}
but even this problem might be challenging. We thus replace it by the linearized version
\begin{multline}
\label{equa:approximate:argmin}
\mu_t = \argmin_{\mu\in M}
\Biggl\{
\sum_{s=1}^{t-1}\left< \mu ,\nabla_{\mu=\mu_s} \mathbb{E}_{\theta\sim q_{\mu} }[\ell_s(\theta)] \right>
\\
+ \frac{ D_\phi(q_\mu||\pi) }{\eta}
\Biggr\}.
\end{multline}
Observe that when $\mu\mapsto \mathbb{E}_{\theta\sim q_{\mu_s} }[\ell_s(\theta)]$ is convex,~\eqref{equa:approximate:argmin} can be seen as a convex relaxation of~\eqref{equa:approximate:argmin:constaint} as
$ \mathbb{E}_{\theta\sim q_{\mu} }[\ell_s(\theta)] \leq \mathbb{E}_{\theta\sim q_{\mu_s} }[\ell_s(\theta)] + \left< \mu ,\nabla_{\mu=\mu_s} \mathbb{E}_{\theta\sim q_{\mu} }[\ell_s(\theta)] \right>$.

\begin{prop}
\label{prop:posterior:general:closedform:approximate}
 Let $F(\mu) = D_\phi(q_\mu||\pi)$. Assume that $F$ is a differentiable and strictly convex function on $\mathbb{R}^d$, then $F^*$ is differentiable with
 \begin{equation}
 \nabla F^*(\lambda) = \argmax_{\mu\in M} \left[\left< \mu,\lambda\right>  - F(\mu) \right].
 \end{equation}
 Then the solution of~\eqref{equa:approximate:argmin} exists, is unique and given by
 \begin{equation}
\mu_t =  \nabla F^* \left( - \eta \sum_{s=1}^{t-1} \nabla_{\mu=\mu_s} \mathbb{E}_{\theta\sim q_{\mu} }[\ell_s(\theta)] \right).
\end{equation}
\end{prop}

Note the ``Mirror Descent'' structure of this strategy: we can simply initialize $\lambda_0=0$, and update at each step:
\begin{equation}
 \left\{
 \begin{array}{l}
  \lambda_t = \lambda_{t-1} -\eta \nabla_{\mu=\mu_{t-1}} \mathbb{E}_{\theta\sim q_{\mu} }[\ell_{t-1}(\theta)] ,
  \\
  \mu_t =  \nabla F^* \left( \lambda_t \right)
 \end{array}
 \right.
\end{equation}
(on mirror descent, see~\cite{mirrorD}, and~\cite{shalev2012online} for an analysis in the online setting).
That is, we have a simple update rule for the ``dual parameters'' $\lambda_t$, and then we compute $\mu_t = \nabla F^* ( \lambda_t )$. An anonymous Referee also pointed out a similarity with ``dual averaging''~\cite{xiao2010dual}.

\subsection{Regret bound}

\begin{thm}
\label{th:regret_bound:approx}
Let $\|\cdot\|$ be a norm on $\mathbb{R}^d$. If each $\mu\mapsto \mathbb{E}_{\theta\sim q_\mu}[\ell_s(\theta)] $ is convex and $L$-Lipschitz with respect to $\|\cdot\|$, if $\mu\mapsto D_\phi(q_\mu||\pi)$ is $\alpha$-strongly convex with respect to $\|\cdot\|$,
 \begin{multline}
 \sum_{t=1}^T \mathbb{E}_{\theta\sim q_{\mu_t}}[\ell_t(\theta)]
 \leq \inf_{\mu\in M}
 \Biggl\{
 \sum_{t=1}^T \mathbb{E}_{\theta \sim q_{\mu}}[\ell_t(\theta)]
 \\
 + \frac{\eta L^2 T}{\alpha} + \frac{D_\phi(q_\mu||\pi) }{\eta}
 \Biggr\}.
 \end{multline}
\end{thm}

Let us consider for example a location scale family $(q_{\mu})_{\mu\in\mathbb{M}}$ with $\mu=(m,C)$, $m\in\mathbb{R}^k$ and $C$ is a $k\times k$ matrix. That is, when $\vartheta \sim q_{(0,I_k)}$, then $m+C\vartheta \sim q_{(m,C)}$. It is proven in~\cite{ConvexDomke}, under minimal assumptions on $q_{(0,I_k)}$, that if $\theta\mapsto\ell_t(\theta)$ is convex, then so is $(m,C)\mapsto \mathbb{E}_{\theta\sim q_{(m,C)}}[\ell_t(\theta)]$. In~\cite{cherief2019generalization}, it is proven that if $\theta\mapsto\ell_t(\theta)$ is $L$-Lipschitz, then $(m,C)\mapsto \mathbb{E}_{\theta\sim q_{(m,C)}}[\ell_t(\theta)]$ is $2L$-Lipschitz.

\begin{exm}
Consider Gaussian distributions. Using the above parametrization $\mu=(m,C)$ with $q_\mu = q_{(m,C)} = \mathcal{N}(m,C^T C)$, $C\in UT(d)$ the set of full-rank upper triangular $d\times d$ real matrices, and chosing as a prior $\pi = q_{(\bar{m},\bar{C})}$ we have
\begin{multline}
{\rm KL}(q_{(m,C)},q_{(\bar{m},\bar{C})})
  = \frac{ (m-\bar{m})^T (\bar{C}^T \bar{C})^{-1} (m-\bar{m}) }{2}
  \\
  + \frac{ {\rm tr}[(\bar{C}^T\bar{C})^{-1} (C^T C)] 
  + \log\left(\frac{{\rm det}(\bar{C}^T\bar{C})}{{\rm det}(C^T C)}\right) -d }{2}
\end{multline}
which is known to be strongly convex on $\mathbb{R}^d \times \mathcal{M}_C$ where $\mathcal{M}_C$ is any closed bounded subset of $UT(d)$. Formulas for the updates are derived in~\cite{cherief2019generalization}.
\end{exm}

Other parametrizations can also be used in practice. For exponential families,~\cite{khan2018} proposed a parametrization based on the expectation of the sufficient statistics. It enjoys very nice properties, and leads to excellent results in practice. However, Theorem~\ref{th:regret_bound:approx} cannot be applied as the convexity assumption is generally not satisfied with this parametrization. The analysis of this algorithm in this case remains an important open question.

\section{Proofs}
\label{section:proofs}

We first remind a classical result in convex analysis, e.g page 95 in~\cite{Boyd} or (2.13) page 43 in~\cite{shalev2012online}.

\begin{lemma}
\label{lemma:conjugate:differential}
 Let $f:\mathbb{R}^d \rightarrow \mathbb{R}\cup\{+\infty\} $ be a function that is differentiable and strictly convex. Then, its convex conjugate $f^*$ is differentiable and
 \begin{equation}
  \nabla f^*(y) = \argmax_{x\in\mathbb{R}^d} \left[ x^T y - f(x) \right].
 \end{equation}
\end{lemma}

\noindent {\it Proof of Theorem~\ref{thm:regret_bound}}: Let us start by proving the existence. For the sake of shortness, put
\begin{equation}
 F(\rho) = \sum_{s=1}^{t-1} \int \ell_s(\theta) \rho({\rm d}\theta) +  \frac{D_\phi(\rho||\pi)}{\eta}
\end{equation}
for any $\rho\in\mathcal{P}(\Theta)$ and $C =  \inf_{\rho\in\mathcal{P}(\Theta)} F(\rho)$. For any $n\in\mathbb{N}$ there is a $\rho_n^t $ such that $ C \leq F(\rho_n^t) \leq C + 1/n $.
Also, the $\rho_n^t$ are absolutely continous with respect to $\pi$, otherwise, $D_\phi(\rho^t_n||\pi)=+\infty$. Then
\begin{align*}
 C & \leq F\left(\frac{ \rho^t_n+\rho^t_m }{2}\right)
 \\
 & \leq  \frac{F(\rho^t_n)+F(\rho^t_m)}{2} - \alpha N(\rho^t_n-\rho^t_m)^2/2
 \\
 & \leq C + 1/(2n)+ 1/(2m)- \alpha N(\rho^t_n-\rho^t_m)^2/2
\end{align*}
which leads to $
 N(\rho^t_n-\rho^t_m)^2 \leq 1/(\alpha n) + 1/(\alpha m),
$
proving that $\rho^t_n$ is a Cauchy sequence w.r.t the norm $N$. Thus, it is also a Cauchy sequence w.r.t the norm $N_{{\rm TV}}$ by the inequality $N(\rho^t_n-\rho^t_m)\geq N_{{\rm TV}}(\rho^t_n-\rho^t_m)$. From Proposition A.10 page 512 in~\cite{ghosal2017fundamentals}, the set of probability distributions that are absolutely continuous with respect to $\pi$ is complete for $N_{{\rm TV}}$, so, there is a $\rho^t_\infty$ absolutely continuous with respect to $\pi$ such that $
N_{{\rm TV}}(\rho^t_n-\rho^t_\infty)\xrightarrow[n\rightarrow \infty]{} 0$.
This can be rewritten as
\begin{equation}
\int \left| \frac{{\rm d}\rho_n^t}{{\rm d}\pi}(\theta) - \frac{{\rm d}\rho_\infty^t}{{\rm d}\pi}(\theta) \right| \pi({\rm d}\theta) \xrightarrow[n\rightarrow \infty]{} 0.
\end{equation}
This means that the nonnegative random variable $\frac{{\rm d}\rho_n^t}{{\rm d}\pi}$ converges to the random variable $\frac{{\rm d}\rho_\infty^t}{{\rm d}\pi}$ in $\mathcal{L}_1$, thus it converges in probability, and thus, there exists a subsequence $\frac{{\rm d}\rho_{n_k}^t}{{\rm d}\pi}$ that converges almost surely to $\frac{{\rm d}\rho_\infty^t}{{\rm d}\pi}$. Now, $\phi$ being lower-bounded, we can use Fatou lemma:
\begin{align*}
C
& \leq F( \rho^t_\infty )
 \\
 & = \int \Biggl[ \sum_{s=1}^{t-1} \ell_s(\theta) \frac{{\rm d}\rho_\infty^t}{{\rm d}\pi}(\theta) + \phi\left(\frac{{\rm d}\rho_\infty^t}{{\rm d}\pi}(\theta)\right)\Biggr] \pi({\rm d}\theta)
 \\
 & = \int  \lim\inf_k \Biggl[ \sum_{s=1}^{t-1} \ell_s(\theta) \frac{{\rm d}\rho_{n_k}^t}{{\rm d}\pi}(\theta) + \phi\left(\frac{{\rm d}\rho_{n_k}^t}{{\rm d}\pi}(\theta)\right)\Biggr] \pi({\rm d}\theta)
 \\
 & \leq \lim\inf_k \int \Biggl[ \sum_{s=1}^{t-1} \ell_s(\theta) \frac{{\rm d}\rho_{n_k}^t}{{\rm d}\pi}(\theta) + \phi\left(\frac{{\rm d}\rho_{n_k}^t}{{\rm d}\pi}(\theta)\right) \Biggr] \pi({\rm d}\theta)
 \\
 & = \lim\inf_k  F(\rho^t_{n_k}) \leq \lim\inf_k \left( C+\frac{1}{n_k} \right) = C
\end{align*}
which proves that $\rho_\infty^t$ is indeed a minimizer of~\eqref{equa:bayes:argmin:general:divergence} (the previous series of inequalities follows the proof of the fact that $\phi$-divergences are lower semi-continuous in Chapter 2 in~\cite{keziou2003utilisation}). Let us now prove its uniqueness: assume that $\tilde{\rho}_\infty^t \neq \rho_\infty^t$ is another minimizer. Put $\bar{\rho}_\infty^t = (\tilde{\rho}_\infty^t+\rho_\infty^t)/2$, using~\eqref{asm:strCvx:D} we have:
\begin{align}
\nonumber
C & \leq F(\bar{\rho}^t_\infty)
 \leq \frac{F(\tilde{\rho}_\infty^t)+F(\rho_\infty^t)}{2}
  - \frac{\alpha}{2}N(\tilde{\rho}_\infty^t-\rho_\infty^t)
  \\
  & = C - \alpha N(\tilde{\rho}_\infty^t-\rho_\infty^t)/2 < C,
\label{equa:proof:unicity}
\end{align}
a contradiction. Thus, $\rho^t = \rho_\infty^t$ exists and is unique.

Let us now prove the regret bound. We follow the main steps of the analysis of the FTRL. We start by proving by induction on $T$ that
\begin{multline}
 \label{equa:proof:recursion}
 \sum_{s=1}^{T} \int \ell_s(\theta) \rho^{s+1}({\rm d}\theta) 
 \\
 \leq \inf_{\rho\in\mathcal{P}(\Theta)} \left[ \sum_{s=1}^{T} \int \ell_s(\theta) \rho({\rm d}\theta) + \frac{D_\phi(\rho||\pi)}{\eta} \right].
\end{multline}
Indeed, for $T=0$, the statement is simply $D_\phi(\rho||\pi)/\eta \geq 0$ that is true by definition of a divergence. Now, assuming that~\eqref{equa:proof:recursion} is true at step $T$, we add $\int \ell_s(\theta) \rho^{T+1}({\rm d}\theta) $ to each side of~\eqref{equa:proof:recursion} to obtain
\begin{multline}
 \label{equa:proof:recursion:2}
 \sum_{s=1}^{T+1} \int \ell_s(\theta) \rho^{s+1}({\rm d}\theta) 
 \leq \int \ell_{T+1}(\theta) \rho^{T+1}({\rm d}\theta)
 \\
 +  \min_{\rho\in\mathcal{P}(\Theta)} \left[ \sum_{s=1}^{T} \int \ell_s(\theta) \rho({\rm d}\theta) +
 \frac{D_\phi(\rho||\pi)}{\eta}\right].
\end{multline}
Upper bounding the minimum in $\rho$ by the value for $\rho=\rho^{T+1}$ we obtain
\begin{align}
 \label{equa:proof:recursion:3}
 \nonumber
 \sum_{s=1}^{T+1} & \int \ell_s(\theta) \rho^{s+1}({\rm d}\theta)
 \\
 & \leq  \sum_{s=1}^{T+1} \int \ell_s(\theta) \rho^{T+1}({\rm d}\theta) + \frac{D_\phi(\rho^{T+1}||\pi)}{\eta}
 \\
 \label{equa:proof:recursion:4}
& 
 = \min_{\rho \in\mathcal{P}(\Theta)}  \left[ \sum_{s=1}^{T+1} \int \ell_s(\theta) \rho({\rm d}\theta) + \frac{D_\phi(\rho||\pi)}{\eta} \right]
\end{align}
by the definition of $\rho^{T+1}$. This ends the proof of~\eqref{equa:proof:recursion}.

Now that~\eqref{equa:proof:recursion} is proven, adding $\sum_{s=1}^T\int \ell_s(\theta) \rho^s({\rm d}\theta) $ to each side and rearranging the terms leads to
\begin{multline}
 \sum_{s=1}^{T} \int \ell_s(\theta) \rho^s({\rm d}\theta)
  \leq \inf_{\rho\in\mathcal{P}(\Theta)} \Biggl[ \sum_{s=1}^{T} \int \ell_s(\theta) \rho({\rm d}\theta)
  \\
  +  \sum_{s=1}^{T} \left(\int \ell_s(\theta) \rho^s({\rm d}\theta) -  \int \ell_s(\theta) \rho^{s+1}({\rm d}\theta)\right)
  \\
  + \frac{D_\phi(\rho||\pi)}{\eta} \Biggr].
\end{multline}

The last step is thus to prove that, for any $s$,
\begin{equation}
\label{eq:proof:increment}
 \int \ell_s(\theta) \rho^s({\rm d}\theta) -  \int \ell_s(\theta) \rho^{s+1}({\rm d}\theta) \leq \frac{\eta L^2}{\alpha}.
\end{equation}

First, by~\eqref{asm:Lip:ell},
\begin{multline}
\label{eq:proof:Lip}
 \int \ell_s(\theta) \rho^s({\rm d}\theta) -  \int \ell_s(\theta) \rho^{s+1}({\rm d}\theta) \\
 \leq L N (\rho^s-\rho^{s+1}).
\end{multline}
Define $H_s(\rho) = \int \sum_{t=1}^{s-1} \ell_t(\theta) \rho({\rm d}\theta) + D_\phi(\rho||\pi)/\eta$. Dividing~\eqref{asm:strCvx:D} by $\eta$ and adding $\gamma \int \sum_{t=1}^{s-1} \ell_t(\theta) \rho({\rm d}\theta) + (1-\gamma)\int \sum_{t=1}^{s-1} \ell_s(\theta) \rho'({\rm d}\theta)$ to each side, we obtain:
 \begin{multline}
  \label{eqn:strCvx:Hs}
  H_s(\gamma \rho + (1-\gamma)\rho ') 
  \leq \gamma H_s(\rho) + (1-\gamma)H_s(\rho') \\
  - \frac{2 \alpha}{\eta} \gamma (1-\gamma) N(\rho-\rho')^2 .
\end{multline}
Now, put $ h_s(u) = H_s(u\rho^{s}+(1-u)\rho^{s+1}) $.
Thanks to~\eqref{eqn:strCvx:Hs}, we have $
 h_s(\gamma u + (1-\gamma)u') \leq \gamma h_s(u) + (1-\gamma)h_s(u') -\frac{\alpha}{2}\gamma(1-\gamma)(u-u')^2 N(\rho^s - \rho^{s+1})^2 $, that is: $h_s$ is $\alpha N(\rho^s - \rho^{s+1})^2$-strongly convex. Moreover, $h_s(u)$ is minimized by $u=0$, because by definition, $H_s(\rho)$ is minimized by $\rho = \rho^s$. Using the well-known property of strongly convex functions of a real variable, we obtain:
\begin{align}
 h_s(u) \geq h_s(0) + \frac{\alpha N(\rho^{s}-\rho^{s+1})^2 }{2\eta} u^2
\end{align}
and so, for $u=1$,
\begin{align}
\label{eqn:strCvx:Hs:1}
 H_s(\rho_{s})\geq H_s(\rho^{s+1}) + \frac{\alpha N(\rho_{s}-\rho^{s+1})^2 }{2\eta}.
\end{align}
We obtain in a similar way:
\begin{align}
\label{eqn:strCvx:Hs:2}
 H_{s+1}(\rho^{s+1})\geq H_{s+1}(\rho^{s}) + \frac{\alpha N(\rho_{s}-\rho^{s+1})^2 }{2\eta}.
\end{align}
Summing~\eqref{eqn:strCvx:Hs:1} and~\eqref{eqn:strCvx:Hs:2} gives:
\begin{multline}
\label{eq:proof:strCvx}
 \int \ell_s(\theta) \rho^s({\rm d}\theta) -  \int \ell_s(\theta) \rho^{s+1}({\rm d}\theta) \\
 \geq \frac{\alpha N(\rho^s-\rho^{s+1})^2}{\eta}.
\end{multline}
Combining~\eqref{eq:proof:strCvx} with~\eqref{eq:proof:Lip} gives:
\begin{multline}
  N(\rho^s-\rho^{s+1}) \\
  \leq \sqrt{\frac{\eta}{\alpha} \left( \int \ell_s(\theta) \rho^s({\rm d}\theta) -  \int \ell_s(\theta) \rho^{s+1}({\rm d}\theta)\right) }
\end{multline}
which, using again~\eqref{eq:proof:Lip}, gives~\eqref{eq:proof:increment}.
$\square$

\bigskip

\noindent {\it Proof of Proposition~\ref{prop:posterior:general:closedform}}: First, note that~\eqref{equa:prop:tilde} is obvious from the definition of $\tilde{\phi}$. Then apply Lemma~\ref{lemma:conjugate:differential} to $f=\tilde{\phi}$ that is $\alpha$-strongly convex. We obtain~\eqref{equa:posterior:general:differential}.

Let us now define $  F_{\phi,\pi}(\rho) =  D_\phi(\rho||\pi)$
and its convex conjugate, for $g:\Theta \rightarrow \mathbb{R}$ that is $\pi$-integrable,
\begin{equation}
 F^*_{\phi,\pi}(g) = \sup_{\rho\in\mathcal{P}_{D,\phi}(\Theta)} \left[ \int g(\theta) \rho({\rm d}\theta) - D_\phi(\rho||\pi)\right].
\end{equation}
Then, by Proposition 4.3.2 in~\cite{agrawal2020optimal},
\begin{equation}
\label{equa:agrawal}
 F^*_{\phi,\pi}(g) = \inf_{\lambda\in\mathbb{R}} \left\{ \int \tilde{\phi}^*(g(\theta)+\lambda) \pi({\rm d}\theta) -\lambda \right\},
\end{equation}
where the infimum is actually reached as soon as it is finite.

In our case, we apply this result to the nonpositive, $\pi$-integrable function
\begin{equation}
 g_t(\theta) = -\eta \sum_{s=1}^{t-1}\ell_s(\theta).
\end{equation}
Using Jensen's inequality, we have:
\begin{multline}
\int \tilde{\phi}^*(g_t(\theta)+\lambda) \pi({\rm d}\theta) -\lambda
\\
\geq 
 \tilde{\phi}^*\left(\int g_t(\theta) \pi({\rm d}\theta)+\lambda\right) -\lambda.
\end{multline}
This quantity is convex, $\geq 0$ when $\lambda\leq 0$, and $\rightarrow \infty$ when $\lambda\rightarrow \infty$. So, its infimum is finite, and thus, according to~\cite{agrawal2020optimal}, it is reached by some $\lambda = \lambda_t$.

Let us now define $\rho^t$ as in~\eqref{equa:posterior:general:closedform:KL}:
$ \rho^t({\rm d}\theta) = \nabla \tilde{\phi}^*(\lambda_t+g_t(\theta)) \pi({\rm d}\theta)$.
A first step is to check that $\rho^t$ is indeed a probability distribution. By differentiating~\eqref{equa:posterior:general:constant:closedform} with respect to $\lambda$ we obtain:
\begin{equation}
 \frac{\partial}{\partial \lambda} \left[ \int \tilde{\phi}^*(\lambda_t+g_t(\theta)) \pi({\rm d}\theta)\right]_{\lambda=\lambda_t} = 1.
\end{equation}
Note that $\nabla \tilde{\phi}^*$ is the differential of a convex, differentiable function. Thus, it is a nondecreasing function, and it has no jumps. So, it is continuous, and so, we have
\begin{equation}
  \int \rho^t ({\rm d}\theta) = \int \nabla \tilde{\phi}^*(\lambda+g(\theta)) \pi({\rm d}\theta)=1.
\end{equation}

Let us now remind the following formula, which can be found for example in~\cite{Boyd} page 95, for a convex and differentiable function $f$:
\begin{equation}
 f^*(\nabla f(x)) = x^T \nabla f(x) - f(x).
\end{equation}
Applying this formula to $f=\tilde{\phi}^*$ that is convex and differentiable, we obtain:
\begin{equation}
 \tilde{\phi}^{**}(\nabla \tilde{\phi}^*(x)) = x^T \nabla \tilde{\phi}^*(x) - \tilde{\phi}^*(x).
\end{equation}
Now, it is easy to check that the function $\tilde{\phi}$ is upper semicontinuous and convex. So, $\tilde{\phi}^{**}=\tilde{\phi}$ (e.g Exercice 3.39 page 121 in~\cite{Boyd}), and we obtain:
\begin{equation}
\label{equa:proof:boyd}
 \tilde{\phi}(\nabla \tilde{\phi}^*(x)) = x^T \nabla \tilde{\phi}^*(x) - \tilde{\phi}^*(x).
\end{equation}
So, we have:
\begin{multline*}
\int \left[- \frac{ g_t(\theta)}{\eta} \right] \rho^t({\rm d}\theta) + \frac{D_\phi(\rho^t||\pi)}{\eta}
\\
 = \int \Biggl[ -\frac{ g_t(\theta)}{\eta} \nabla \tilde{\phi}^*(\lambda_t+g_t(\theta))  
 \\
 + \frac{1}{\eta}\tilde{\phi}\left( \nabla \tilde{\phi}^*(\lambda_t+g_t(\theta)) \right)\Biggr] \pi({\rm d}\theta)
\end{multline*}
and applying~\eqref{equa:proof:boyd} gives
\begin{align}
\nonumber
 \int & \left[- \frac{ g_t(\theta)}{\eta} \right] \rho^t({\rm d}\theta) + \frac{D_\phi(\rho^t||\pi)}{\eta}
\\
\nonumber
& = \int \Biggl[ -\frac{ g_t(\theta)}{\eta} \nabla \tilde{\phi}^*(\lambda_t+g(\theta))
\\
\nonumber
& \quad \quad \quad \quad
+ \frac{(\lambda_t+g(\theta))}{\eta} \nabla \tilde{\phi}^*(\lambda_t+g(\theta)) 
\\
\nonumber
& \quad \quad \quad \quad
- \tilde{\phi}^*(\lambda_t + g_t(\theta)) \Biggr] \pi({\rm d}\theta)
\\
\nonumber
& = \lambda_t - \int \tilde{\phi}^*(\lambda_t + g_t(\theta)) \pi({\rm d}\theta)
\\
& = \min_{\rho \in\mathcal{P}_{D,\pi}(\Theta)} \left[-\int   \frac{ g_t(\theta)}{\eta}  \rho^t({\rm d}\theta) + \frac{D_\phi(\rho^t||\pi)}{\eta} \right]
\end{align}
by~\eqref{equa:agrawal}. So $\rho^t$ minimizes the desired criterion.
$\square$

\bigskip

The proof of Proposition~\ref{prop:posterior:general:closedform:approximate} and Theorem~\ref{th:regret_bound:approx} are provided in Appendix B.

\section*{Acknowledgements}
 
The anonymous Referees suggested many clarifications and connections to existing works. I thank them deeply for this. I also would like to thank Emtiyaz Khan and all the members of the ABI team (RIKEN AIP), Dimitri Meunier (IIT Genoa), Badr-Eddine Ch\'erief-Abdellatif (Univ. of Oxford), Jeremias Knoblauch and Lionel Riou-Durand (Univ. of Warwick) for useful discussions/comments that led to improvements of the paper.

\newpage

\quad

\newpage

\appendix

\begin{center}
 {\bf APPENDIX}
\end{center}

We remind that Appendix A contains an application of Theorem~\eqref{thm:regret_bound} to the continuous case. Appendix B contains the proofs of Proposition~\ref{prop:posterior:general:closedform:approximate} and Theorem~\ref{th:regret_bound:approx}. Finally, Appendix C contains the derivation of  $\rho^t$ in the finite case thanks to Lagrange method of multipliers.

\section{Comparison of the Bounds in the Continuous Case}

 As another example of application of Theorem~\eqref{thm:regret_bound}, let us consider the case $\ell_t(\theta)=(\theta-y_t)^2$. We assume that $\sup_{t\in\mathbb{N}}|y_t|=C<+\infty$. We prove the following statements:
 \begin{itemize}
 \item for some choice of $\eta$ and $\pi$, EWA (that is,~\eqref{equa:bayes:argmin:general:divergence} with $D_\phi={\rm KL}$) leads to
  \begin{align*}
 \sum_{t=1}^T \mathbb{E}_{\theta\sim \rho^t}[\ell_t(\theta)]
 & \leq \inf_{m\in[-C,C]}\Biggl\{ \sum_{t=1}^T (y_t-m)^2
 \\
 & \quad \quad + 4 C^2 \sqrt{ T \log(T)} (1+o(1))
 \\
 & = \sum_{t=1}^T \left(y_t- \bar{y}_T\right)^2
 \\
 & \quad \quad + 4 C^2 \sqrt{ T \log(T)} (1+o(1))
 \Biggr\}
 \end{align*}
 where $\bar{y}_T = (1/T) \sum_{t=1}^T y_t$, but $C$ has to be known by the user to reach this.
 \item for some choice of $\eta$ and $\pi$,~\eqref{equa:bayes:argmin:general:divergence} with $D_\phi=\chi^2$ leads to
\begin{multline}
\label{bound:continuous:chi2}
 \sum_{t=1}^T \mathbb{E}_{\theta\sim \rho^t}[\ell_t(\theta)]
 \\
 \leq \inf_{m\in\mathbb{R}}
 \Biggl\{
 \sum_{t=1}^T (m-y_t)^2 + C' T^{\frac{2}{3}}(1+|m|)^{5}
 \Biggr\}
\end{multline}
where $C'$ is a constant that depends only on $C$, and none of these constants have to be known by the user.
\end{itemize}

 There are various ways of using EWA in this context. The important point is that they all require the support of the prior $\pi$ to be bounded (or to truncate its support at some point):
\begin{enumerate}
\item a first option is to use as a prior $\pi$ the uniform distribution on $[-C,C]$. Of course, this is possible only if one knows $C$ in advance! In this case, the losses are bounded by $4C^2$ and so the regret bound is given by
 \begin{multline}
 \sum_{t=1}^T \mathbb{E}_{\theta\sim \rho^t}[\ell_t(\theta)]
 \leq \inf_{\rho\in\mathcal{P}(\Theta)}
 \Biggl\{
 \sum_{t=1}^T \mathbb{E}_{\theta \sim \rho}[(y_t-\theta)^2]
 \\
 + \frac{\eta C^2 T}{2} + \frac{ {\rm KL}(\rho||\pi) }{\eta}
 \Biggr\}.
 \end{multline} 
 For $m\in[-C,C]$ and $\delta\in(0,1)$, define $\rho_{m,\delta}$ as the uniform distribution on an inverval of length $\delta C$ that contains $m$ (one could think of $[m-\delta C/2,m+\delta C/2]$ but when $m=C$, this interval would not be included in $[-C,C]$...). We obtain:
 \begin{align*}
 \sum_{t=1}^T & \mathbb{E}_{\theta\sim \rho^t}[\ell_t(\theta)]
 \\
 & \leq \inf_{m\in[-C,C]} \inf_{\delta\in(0,1)}
 \Biggl\{
 \sum_{t=1}^T \mathbb{E}_{\theta \sim \rho}[(\theta-y_t)^2]
 \\
 & \quad \quad
 + 8 \eta C^4 T + \frac{ \log\left(\frac{2}{\delta}\right) }{\eta}
 \Biggr\}
 \\
 \nonumber
 & \leq \inf_{m\in[-C,C]} \inf_{\delta\in(0,1)}\Biggl\{ \sum_{t=1}^T \Bigl( (y_t-m)^2 + C^2\delta^2
 \\
 & \quad \quad 
 + 2C\delta|y_t-m| \Bigr) + 8 \eta C^4 T + \frac{ \log\left(\frac{2}{\delta}\right) }{\eta}
 \Biggr\}
 \\
 \nonumber
 & \leq \inf_{m\in[-C,C]} \inf_{\delta\in(0,1)} \Biggl\{ \sum_{t=1}^T (y_t-m)^2
 +  5TC^2 \delta
 \\
 & \quad \quad
 + 8 \eta C^4 T + \frac{ \log\left(\frac{2}{\delta}\right) }{\eta}
 \Biggr\}
  \\
 & = \inf_{m\in[-C,C]} \Biggl\{ \sum_{t=1}^T (y_t-m)^2
 \\
 & \quad \quad
 + 8 \eta C^4 T + \frac{ 1+\log\left(10 T C^2 \eta\right) }{\eta}
 \Biggr\}
 \end{align*}
 reached for $\delta = 1/(5 \eta T C^2)$ (in $(0,1)$ for $T$ large enough). The choice $\eta = \sqrt{\log(T)}/(4C^2 \sqrt{T}) $ gives:
 \begin{multline}
 \sum_{t=1}^T \mathbb{E}_{\theta\sim \rho^t}[\ell_t(\theta)]
 \leq \inf_{m\in[-C,C]}\Biggl\{ \sum_{t=1}^T (y_t-m)^2
 \\
 + 4 C^2 \sqrt{ T \log(T)} (1+o(1))
 \Biggr\}.
 \label{appendix:oracle:EWA1}
 \end{multline}
 \item a second strategy is detailed for example in~\cite{gerchinovitz2011prediction}, it consists in taking a heavy-tailed distribution on $\mathbb{R}$ for $\pi$, but to use as a predictor the projection of $\theta$ on the interval $[-C,C]$, that is, changing the loss in $|y_t - {\rm proj}_{[-C,C]}(\theta)|$. This would lead to a regret bound similar to~\eqref{appendix:oracle:EWA1}, without improving the applicability of the result, in the sense that one has to know $C$ to use the procedure.
 \item a third approach was mentioned by an anonymous Referee, and can in principle be used when $C$ is unknown. In this case, we take $\pi$ as the uniform prior on $[-c,c]$ for some $c>0$. The important point is that the loss $\ell_t(\theta)$ belongs to the interval $[0,(c+C)^2]$ whose upper bound is unknown. Based on techniques developped in~\cite{cesa2001worst,auer2002nonstochastic}, Theorem 6 in~\cite{cesa2007improved} upper bounds the regret of an adaptive version of EWA that can be used in the case where the losses belongs to an (unknown) bounded interval. This theorem is written in the finite $\Theta$ case, but it seems to be direct to extend the result to the general case. If one is ``lucky'', that is, if $c\geq C$, then one would recover a regret bound similar to~\eqref{appendix:oracle:EWA1}. However, it might be that $C>c$. In this case, we only have the guarantee to perform as well as the best predictor in $[-c,c]$. If the best predictor $m$ satisfies $|m|>c$ then this gives a linear regret.
\end{enumerate}
 
 Let us now use the strategy~\eqref{equa:bayes:argmin:general:divergence} with $D$ being the $\chi^2$ divergence, and with a prior $\pi$ that is the student distribution $\mathcal{T}(k)$ with $k=4$ degrees of freedom. First,
 \begin{multline}
   \int \ell_t(\theta)^2  \pi({\rm d}\theta) =  \int |y_t-\theta|^4  \pi({\rm d}\theta) 
   \\
   \leq 8 \int |y_t|^4  \pi({\rm d}\theta) + 8 \int |\theta|^4  \pi({\rm d}\theta) \leq 8(C^4+24).
 \end{multline}
This gives the regret bound
  \begin{multline}
 \sum_{t=1}^T \mathbb{E}_{\theta\sim \rho^t}[\ell_t(\theta)]
 \leq \inf_{\rho\in\mathcal{P}(\Theta)}
 \Biggl\{
 \sum_{t=1}^T \mathbb{E}_{\theta \sim \rho}[(y_t-\theta)^2]
 \\
 + \eta 8 (C^4+24)T + \frac{ \chi^2(\rho||\pi) }{\eta}
 \Biggr\}
 \end{multline}
 and here, let us consider, for $m\in\mathbb{R}$ and $\delta\in(0,1)$, the uniform distribution $\rho_{m,\delta}$ on an interval of length $\delta C$ that contains $m$. The regret bound becomes:
 \begin{multline}
 \sum_{t=1}^T \mathbb{E}_{\theta\sim \rho^t}[\ell_t(\theta)]
 \\
 \leq \inf_{m\in\mathbb{R}} \inf_{\delta\in(0,1)}
 \Biggl\{
 \sum_{t=1}^T (m-y_t)^2 + \delta 5 C^2 T
 \\
 + \eta 8 (C^4+24)T + \frac{ \chi^2(\rho_{m,\delta}||\pi) }{\eta}
 \Biggr\}.
\end{multline}
Note that the density of $\mathcal{T}(k)$ with respect to the Lebesgue measure is given by:
\begin{equation}
 \frac{1}{\sqrt{k\pi}}\frac{\Gamma\left(\frac{k+1}{2}\right)}{\Gamma\left(\frac{k}{2}\right)} \left(1+\frac{t^2}{k}\right)^{-\frac{k+1}{2}}
\end{equation}
so we can derive the upper bound
\begin{equation}
 \chi^2(\rho_{m,\delta}||\pi) \leq
 \frac{C''}{\delta \eta} (1+|m|)^{5}
\end{equation}
for some $C''>0$ that depends only on $C$. This time, the choices $\delta = \eta = 1/T^{1/3}$ lead to
\begin{multline}
 \sum_{t=1}^T \mathbb{E}_{\theta\sim \rho^t}[\ell_t(\theta)]
 \\
 \leq \inf_{m\in\mathbb{R}}
 \Biggl\{
 \sum_{t=1}^T (m-y_t)^2 + C' T^{\frac{2}{3}}(1+|m|)^{5}
 \Biggr\}
\end{multline}
where $C'$ is a constant that depends only on $C$. The important point is that the strategy can be implemented without the knowledge of $C$ nor $C'$. But also, this has an important cost, that is, the regret is now in $T^{2/3}$.

 \begin{rmk}
  An anonymous Referee suggested that it is possible to first build predictors on nested intervals, and then to aggregate them via EWA to obtain adaptation to the unknown constant $C$. However, these predictors are not uniformly bounded, so we don't see how to apply the standard results on EWA to them.
  
  However, this suggestion leads to an improvement on~\eqref{bound:continuous:chi2} that will combine the ideas of EWA and non-exponentially weighted aggregation. The idea is to use EWA on nested intervals, and then to aggregate them using the $\chi^2$ bound, which does not require boundedness.
  
  More precisely, define $\rho_k^t$ as the result of using EWA with a uniform prior on $[-k,k]$, for any $k\in\mathbb{N}\setminus\{0\}$. We have:
    \begin{align*}
 \sum_{t=1}^T \mathbb{E}_{\theta\sim \rho^t_k}[\ell_t(\theta)]
 & \leq \inf_{m\in[-k,k]}\Biggl\{ \sum_{t=1}^T (y_t-m)^2
 \\
 & \quad \quad + 4 k^2 \sqrt{ T \log(T)} (1+o(1))
 \Biggr\}.
 \end{align*}
 It is then possible to adapt Corollary~\ref{coro:regret:chi2} to aggregate the various $\rho^t_k$, using a prior $\pi$ on $k$. This leads to a posterior $\rho^t$ on $k$ such that
 \begin{align*}
  \sum_{t=1}^T & \mathbb{E}_{k\sim \rho^t} \mathbb{E}_{\theta\sim \rho^t_k}[\ell_t(\theta)]
  \\
  & \leq
  \inf_{k\geq 1} \left\{  \sum_{t=1}^T \mathbb{E}_{\theta\sim \rho^t_k}[\ell_t(\theta)] + \frac{\eta L^2 T}{2} + \frac{\frac{1}{\pi(k) -1}}{\eta} \right\}
  \\
  & \leq   \inf_{k\geq 1}  \inf_{m\in[-k,k]}\Biggl\{ \sum_{t=1}^T (y_t-m)^2
 \\
 & \quad  + 4 k^2 \sqrt{ T \log(T)} (1+o(1)) + \frac{\eta L^2 T}{2} + \frac{\frac{1}{\pi(k)}-1}{\eta} \Biggr\}
 \end{align*}
 where
 $$ L^2 = 2C^2 + 2 \sum_{k=1}^\infty \pi(k) k^2 .$$
The choice $\pi(k) \propto 1 / k^4 $ and $\eta \propto 1/\sqrt{T}$ leads to a bound in:
 \begin{multline*}
  \sum_{t=1}^T  \mathbb{E}_{k\sim \rho^t} \mathbb{E}_{\theta\sim \rho^t_k}[\ell_t(\theta)]
 \leq   \inf_{m\in\mathbb{R}}\Biggl\{ \sum_{t=1}^T (y_t-m)^2
 \\
 + (1+m^4+C^2)\sqrt{T\log(T)}(1+o(1)) \Biggr\}
 \end{multline*}
 which improves on~\eqref{bound:continuous:chi2}.
  \end{rmk}

\section{Proofs of the Results in Section~\ref{section:approximate}}

\noindent {\it Proof of Proposition~\ref{prop:posterior:general:closedform:approximate}}: It is a direct application of Lemma~\ref{lemma:conjugate:differential} to $f=F$.
$\square$

\bigskip

\noindent {\it Proof of Theorem~\ref{th:regret_bound:approx}}: This proof follows step by step the classical analysis of FTRL, but we provide it for the sake of completeness. For short, let $\bar{L}_t(\mu) := \mathbb{E}_{\theta\sim q_\mu}[\ell_t(\theta)]$.
First, by assumption, $\bar{L}_t$ is convex. By definition of the subgradient of a convex function,
\begin{align}
\sum_{t=1}^{T} & \mathbb{E}_{\theta\sim q_{\mu_t}}[ \ell_t(\theta) ]
-  \sum_{t=1}^{T} \mathbb{E}_{\theta\sim q_\mu} [\ell_t(\theta)] 
\nonumber
\\
& = \sum_{t=1}^T \bar{L}_t(\mu_t) - \sum_{t=1}^t \bar{L}_t(\mu)
\nonumber
\\
& \leq \sum_{t=1}^T \mu_t^T \nabla \bar{L}_t(\mu_t) - \sum_{t=1}^T \mu^T \nabla \bar{L}_t(\mu_t).
\label{eq-step1}
\end{align}

Then, we prove by recursion on $T$ that for any $\mu\in\mathbb{R}^d$,
\begin{multline}
\sum_{t=1}^T \mu_t^T \nabla \bar{L}_t(\mu_t) - \sum_{t=1}^T \mu^T \nabla \bar{L}_t(\mu_t)
\\
\leq
\sum_{t=1}^T \mu_t^T \nabla \bar{L}_t(\mu_t) - \sum_{t=1}^T \mu_{t+1}^T \nabla \bar{L}_t(\mu_t)
\\
+ \frac{D_\phi(q_\mu||\pi)}{\eta}
\label{eq-lemma23-shalev}
\end{multline}
which is exactly equivalent to
\begin{equation}
 \label{eq-lemma23-shalev-prime}
\sum_{t=1}^T \mu_{t+1}^T \nabla \bar{L}_t(\mu_t)
 \leq  \sum_{t=1}^T \mu^T \nabla \bar{L}_t(\mu_t) + \frac{D_\phi(q_\mu||\pi)}{\eta}.
\end{equation}
Indeed, for $T=0$,~\eqref{eq-lemma23-shalev-prime} just states that $D_\phi(q_\mu||\pi) \geq 0$ which is true by assumption. Assume that~\eqref{eq-lemma23-shalev-prime} holds for some integer $T-1$.
We then have, for all $\mu\in \mathbb{R}^d$,
\begin{align*}
\sum_{t=1}^{T} & \mu_{t+1}^T \nabla \bar{L}_t(\mu_t)
\\
 & = \sum_{t=1}^{T-1} \mu_{t+1}^T \nabla \bar{L}_t(\mu_t) + \mu_{T+1}^T \nabla \bar{L}_T(\mu_T)
 \\
 & \leq  \sum_{t=1}^{T-1} \mu^T \nabla \bar{L}_t(\mu_t) + \frac{D_\phi(q_\mu||\pi)}{\eta} + \mu_{T+1}^T \nabla \bar{L}_T(\mu_T)
\end{align*}
as~\eqref{eq-lemma23-shalev-prime} holds for $T-1$. Apply this to $\mu=\mu_{T+1}$ to get
\begin{align*}
\sum_{t=1}^{T} & \mu_{t+1}^T \nabla \bar{L}_t(\mu_t)
\\
& \leq \sum_{t=1}^{T} \mu_{T+1}^T \nabla \bar{L}_t(\mu_t) + \frac{D(q_{\mu_{T+1}}||\pi)}{\eta}
\\
& = \min_{m\in \mathbb{R}^d} \left[\sum_{t=1}^T m^T \nabla \bar{L}_t(\mu_t) + \frac{D(q_m||\pi)}{\eta}\right]
\\
& \quad \quad (\text{by definition of } \mu_{T+1}),
\\
& \leq \sum_{t=1}^T \mu^T \nabla \bar{L}_t(\mu_t) + \frac{D_\phi(q_\mu||\pi)}{\eta}
\end{align*}
for all $\mu\in \mathbb{R}^d$. Thus,~\eqref{eq-lemma23-shalev-prime} holds for $T$. Thus, by recursion,~\eqref{eq-lemma23-shalev-prime} and~\eqref{eq-lemma23-shalev} hold for all $T\in\mathbb{N}$.

The last step is to prove that for any $t\in\mathbb{N}$,
\begin{equation}
 \label{eq-FTRL}
  \mu_t^T \nabla \bar{L}_t(\mu_t) - \mu_{t+1}^T \nabla \bar{L}_t(\mu_t)
 \leq \frac{\eta L^2}{\alpha}.
\end{equation}
Indeed,
\begin{align}
 \mu_t^T  \nabla \bar{L}_t(\mu_t) - & \mu_{t+1}^T \nabla \bar{L}_t(\mu_t)
 \nonumber
 \\
 & = (\mu_t - \mu_{t+1})^T \nabla \bar{L}_t(\mu_t)
  \nonumber
 \\
 & \leq \|\mu_t-\mu_{t+1}\| \|\nabla \bar{L}_t(\mu_t)\|^*
  \nonumber
 \\
 & \leq L \|\mu_t - \mu_{t+1}\|
 \label{eq-FTRL-step1}
\end{align}
as $\bar{L}_t$ is $L$ Lipschitz w.r.t $\|\cdot\|$ (Lemma 2.6 page 27 in~\cite{shalev2012online} states that the conjugate norm of its gradient is bounded by $L$). Define
$$
G_{t}(\mu) = \sum_{i=1}^{t-1} \mu^T \nabla \bar{L}_i(\mu_i) + \frac{D_\phi(q_\mu||\pi)}{\eta}.
$$
We remind that by assumption, $\mu\mapsto D_\phi(q_\mu||\pi)/\eta$ is $\alpha/\eta$-strongly convex with respect to $\|\cdot\|$. As the sum of a linear function and an $\alpha/\eta$-strongly convex function, $G_t$ is $\alpha/\eta$-strongly convex. So, for any $(\mu,\mu')$,
$$
G_{t}(\mu') - G_{t}(\mu) \geq (\mu'-\mu)^T \nabla G_t(\mu) + \frac{\alpha \|\mu'-\mu\|^2}{2 \eta} .
$$
As a special case, using the fact that $\mu_t$ is a minimizer of $G_t$, we have
$$
G_{t}(\mu_{t+1}) - G_{t}(\mu_t) \geq \frac{\alpha \|\mu_{t+1}-\mu_t\|^2}{2 \eta}.
$$
In the same way,
$$
G_{t+1}(\mu_{t}) - G_{t+1}(\mu_{t+1}) \geq \frac{\alpha \|\mu_{t+1}-\mu_t\|^2}{2 \eta}.
$$
Summing the two previous inequalities gives
$$
\mu_t^T  \nabla \bar{L}_t(\mu_t) - \mu_{t+1}^T \nabla \bar{L}_t(\mu_t) \geq \frac{\alpha \|\mu_{t+1}-\mu_t\|^2}{\eta},
$$
and so, combined with~\eqref{eq-FTRL-step1}, this gives:
\begin{equation*}
 \|\mu_{t+1}-\mu_t\|
  \leq \sqrt{\frac{\eta}{\alpha}\left[ \mu_t^T \nabla \bar{L}_t(\mu_t) - \mu_{t+1}^T \nabla \bar{L}_t(\mu_t)\right]}.
\end{equation*}
Combining this inequality with~\eqref{eq-FTRL-step1} leads to~\eqref{eq-FTRL}.

Plugging~\eqref{eq-step1},~\eqref{eq-lemma23-shalev} and~\eqref{eq-FTRL} together gives
\begin{multline*}
\sum_{t=1}^{T} \mathbb{E}_{\theta\sim q_{\mu_t}}[ \ell_t(\theta) ]
-  \sum_{t=1}^{T} \mathbb{E}_{\theta\sim q_\mu} [\ell_t(\theta)] 
\\
\leq \frac{\eta T L^2}{\alpha} + \frac{D_\phi(q_\mu||\pi)}{\eta},
\end{multline*}
that is the statement of the theorem.
$\square$

\section{Derivation of $\rho^t$ in the Finite Case via Lagrange Method of Multipliers}

Following Remark~\ref{rmk:posterior}, we provide the derivation of $\rho^t$ in the finite case, thanks to Lagrange method of multipliers. We remind that
 \begin{multline}
 \mathcal{L}(\rho^t_1,\dots,\rho^t_M,\lambda,\nu_1,\dots,\nu_M) 
 =  \sum_{j=1}^M \rho^t_j \sum_{s=1}^{t-1} \ell_s(\theta_j) 
 \\
 + \frac{\sum_{j=1}^M \pi_j \phi \left(\frac{\rho^t_j}{\pi_j}\right) }{\eta} + \lambda \frac{ 1-\sum_{j=1}^M \rho^t_j}{\eta} + \sum_{j=1}^M \nu_j \rho^t_j.
 \end{multline}
 So:
 \begin{multline}
 \frac{\partial}{\partial \rho_j^t} \mathcal{L}(\rho^t_1,\dots,\rho^t_M,\lambda,\nu_1,\dots,\nu_M) 
 =  \sum_{s=1}^{t-1} \ell_s(\theta_j) 
 \\
 + \frac{\phi' \left(\frac{\rho^t_j}{\pi_j}\right) }{\eta} +\frac{ -\lambda }{\eta} + \nu_j.
 \end{multline}
 Thus the first-order equation
 \begin{equation}
   \frac{\partial}{\partial \rho_j^t} \mathcal{L}(\rho^t_1,\dots,\rho^t_M,\lambda,\nu_1,\dots,\nu_M) = 0
 \end{equation}
is equivalent to
 \begin{equation}
\phi' \left(\frac{\rho^t_j}{\pi_j}\right) = \lambda -\eta \sum_{s=1}^{t-1} \ell_s(\theta_j) - \eta \nu_j.
 \end{equation}
Intuitively, the next step would be to apply the inverse of the function $\phi'$:
 \begin{equation}
\frac{\rho^t_j}{\pi_j} =  (\phi')^{-1}\left( \lambda -\eta \sum_{s=1}^{t-1} \ell_s(\theta_j) - \eta \nu_j \right).
 \end{equation}
Remind that the first order condition for $\nu_j$ is $\nu_j \geq 0$ and $\nu_j >0 \Leftrightarrow \rho_j = 0$. So, we would obtain the simpler formula:
 \begin{equation}
 \label{appendix:equa:moche}
\rho^t_j = \pi_j \max\left\{ 0, (\phi')^{-1}\left( \lambda -\eta \sum_{s=1}^{t-1} \ell_s(\theta_j) \right)\right\}.
 \end{equation}
It turns out that, under the assumptions of Proposition~\ref{prop:posterior:general:closedform}, $(\phi')^{-1}$ indeed exists and $\nabla \tilde{\phi}^*(y) = \max\{0,(\phi')^{-1}(y) \} $. That is,~\eqref{appendix:equa:moche} is equivalent to~\eqref{equa:posterior:general:closedform}.
\end{document}